\documentclass{segabs}
\usepackage{amsmath}
\usepackage{graphicx}
\usepackage{multicol}
\usepackage{cuted}
\usepackage{subfig}

\begin{document}

\title{Seismic facies recognition based on prestack data using deep convolutional autoencoder}

\renewcommand{\thefootnote}{\fnsymbol{footnote}} 
\renewcommand{\figdir}{Fig} 

\author{Feng Qian$^{1}$, Miao Yin$^{2}$, Ming-Jun Su$^{3}$, Yaojun Wang$^{1}$, Guangmin Hu$^{1}$\\
$^{1}$Center of Information Geoscience, University of Electronic Science and Technology of China, \\
$^{2}$School of Resource and Environment, University of Electronic Science and Technology of China,\\
$^{3}$PetroChina Research Institute of Petroleum Exploration and Development (RIPED) - Northwest, Lanzhou, China
}

\righthead{Seismic facies recognition based on prestack data using deep convolutional autoencoder}

\maketitle

\begin{abstract}
Prestack seismic data carries much useful 
information that can help us find more complex atypical reservoirs. Therefore, 
we are increasingly inclined to use prestack seismic data for seismic 
facies recognition. However, due to the inclusion of excessive redundancy, effective feature extraction from prestack seismic data becomes 
critical. In this paper, we consider seismic facies recognition based on prestack data as an 
image clustering problem in computer vision (CV) by thinking of each prestack 
seismic gather as a picture. 
We propose a convolutional autoencoder (CAE) network for deep feature learning from 
prestack seismic data, which is more effective than principal 
component analysis (PCA) in redundancy removing and valid information extraction. Then, using conventional classification or clustering techniques (e.g. K-means or self-organizing maps) on the extracted features,  
we can achieve seismic facies recognition. We applied our 
method to the prestack data from physical model and LZB region. 
The result shows that 
our approach is superior to the conventionals.

\end{abstract}

\section{Introduction}
Seismic facies recognition is an auxiliary means that we use machine learning to generate facies map which reveals the details of the underlying geological features. 
With the improvement of computer's computing ability, pattern recognition techniques can be used to handle larger seismic data and get more accurate results. 
As a result, seismic facies recognition becomes increasingly significant.
 
So far, almost all supervised and unsupervised classification algorithms including K-means, self-organizing maps (SOM), generative topographic mapping (GTM), support vector machines (SVM) and artificial neural networks (ANN) etc. have been successfully used in seismic facies recognition. 
In fact, the most important parameter in the analysis is not classification method but the correct input attributes \citep{zhao2015comparison}. 
It means that effective feature extraction from seismic data is the key to seismic facies recognition. 
\citet{balz2000fast} developed a methodology to generate a so-called AVO-trace from prestack seismic data which can be classified by conventional clustering techniques. 
\citet{coleou2003unsupervised} extracted features via constructing vectors of dip magnitude, coherence and reflector parallelism. 
\citet{de2006unsupervised} used wavelet transforms to identify seismic trace singularities in each geologically oriented segment.
\citet{gao2007application} applied gray level co-occurrence matrix (GLCM) to generate 
facies map using SOM. 
\citet{roy2013characterizing} proposed an SOM classification workflow based on multiple  seismic attributes. 
Of the seismic facies recognition methods, only a few extract features from prestack seismic data. 

Obviously, prestack seismic data contains richer information, with which we can obtain more details of atypical reservoirs in facies map. 
Nevertheless, there is also a large amount of redundancy in the prestack data that may effect the performance of seismic facies recognition negatively. 
Principal components analysis (PCA) and independent components analysis (ICA) are commonly used to reduce redundancy of attributes to low-dimensional independent meta features \citep{gao2007application}. 
These  approaches just map the original prestack seismic data to a subspace in view of mathematics. 
However, our motivation is to develop a model, which can learn profound features from seismic data automatically. 

In order to construct the model, referring to the latest technologies in the fields of artificial intelligence (AI) and computer vision (CV), we applied convolutional neural network (CNN) in seismic facies recognition. 
Convolutional neural network is one framework of deep learning which has been used effectively in many machine learning tasks. 
With the excellence of feature learning, CNN has achieved particular success in the domain of image classification \citep{krizhevsky2012in}, action recognition \citep{ji20133d}, video classification \citep{karpathy2014large}, speech recognition \citep{abdel2014convolutional} and face recognition \citep{farfade2015multi}. 
 Considering the seismic prestack data as images, we convert seismic facies recognition to image classification problem resolved by using CNN to extract features.

In this paper, we develop a convolutional autoencoder (CAE) network with multiple layers. 
In each layer, higher-level feature maps are generated by convolving features of the lower-level layer with converlutional filters which are trained by minimizing a loss function (explained in the next section). 
Firstly, we input the seismic prestack data to train the network and obtain extracted features. Then we utilize unsupervised pattern recognition techniques to generate the facies map. We have compared the facies maps depicted by our method and other two methods (PCA and using postack data) with two types of prestack seismic data. The result demonstrates the superiority of our approach.

\begin{figure*}[t]
  \centering
  \includegraphics[width=\textwidth]{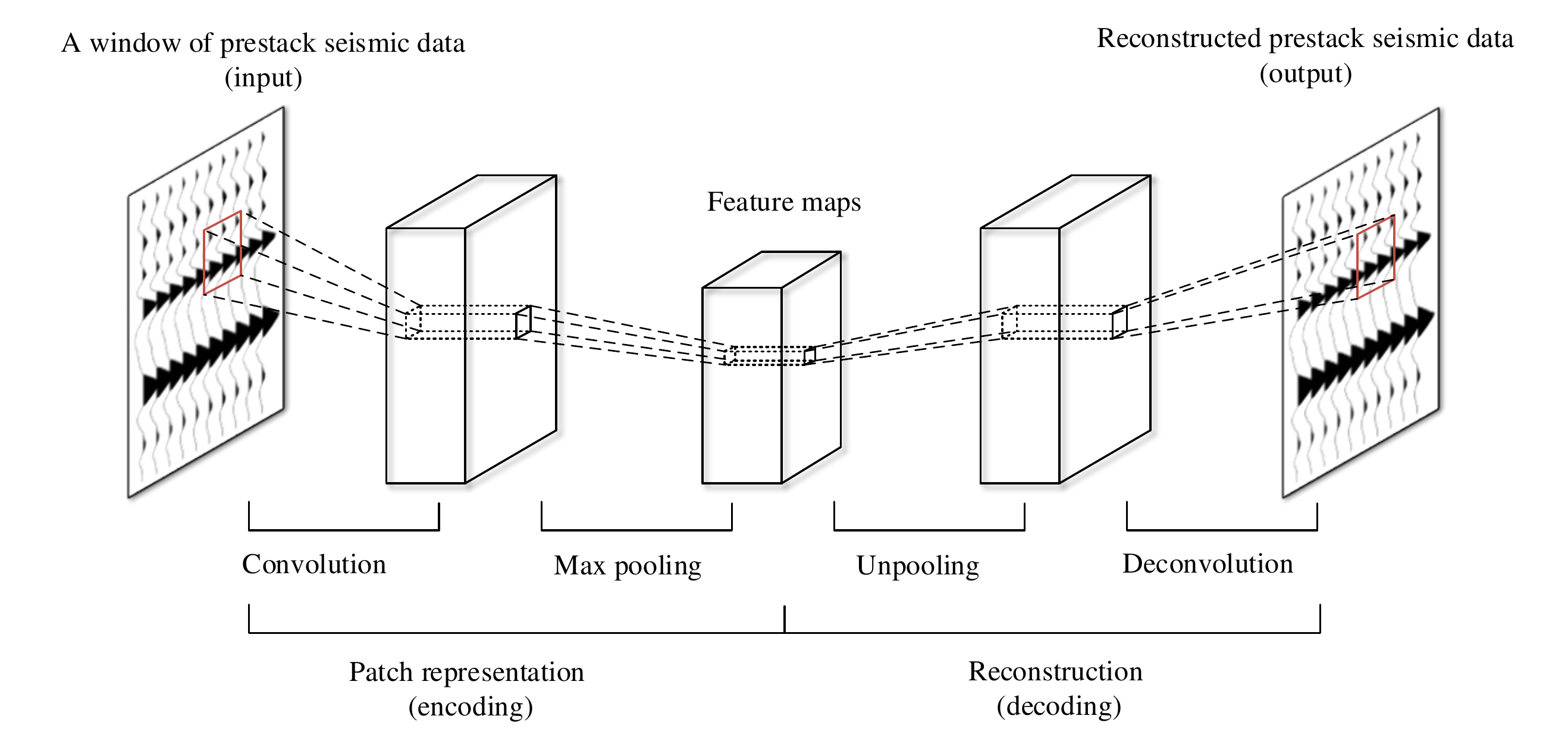}
  \caption{Architecture of convolutional neural network we proposed to learn deep features from prestack seismic data.}
  \label{fig:cnn}
\end{figure*}

\section*{Method}

The workflow of our method mainly consists of the following four steps: \\
(1) Select a proper window of data which contains at least one crest and one trough per trace in prestack seismic gathers along an interpreted horizon; \\
(2) Input these two-dimensional data to the convolutional autoencoder network and train it; \\
(3) Re-input these two-dimensional data to the trained network, obtain features extracted by it;\\
(4) Use unsupervised pattern recognition algorithm to cluster these extracted features and generate facies map. \\

\subsection*{Convolutional Neural Network For Feature Learning}

An overview of our convolutional neural network is given by Figure \ref{fig:cnn}. The input $\mathbf{X}$ is a window of prestack seismic data, which can be considered as an image. Our goal is to learn some features that can best represent $\mathbf{X}$. That means we wish to find a mapping $\Phi$, which makes the reconstruction $\mathbf{Z}$ as similar as possible to $\mathbf{X}$. Feature learning from seismic prestack data involves the following three aspects.

\textbf{Patch representation.} This operation extracts patches from the seismic prestack window data $\mathbf{X}$ and represents each patch as a high-dimensional tensor. 
These tensors comprise a set of feature maps, of which the number is equal to the dimensionality of the tensors. 
At the beginning the input data is convolved, and then a max pooling step is used to shrink the convolutional layers --- taking the maximum of each $2 \times 2$ matrix with a strides of $2$ along both width and height. 
Max pooling reduces the spatial size of the representation and hence to also control overfitting. 
Formally, for a window of prestack data as an input $\mathbf{X}$, the latent representation of the $k^{\text{th}}$ feature map is given by \\
\begin{equation} \label{eq:conv}
  \Phi_{k}(\mathbf{X})= \sigma \left ( \Psi(\mathbf{X} * \mathbf{W}_{k} + \pmb{b}_{k})\right ), 
\end{equation}\\
where $\mathbf{W}$ represents the filter, $\Psi$ donates max pooling, $\pmb{b}$ is the bias which is broadcasted to the whole map. $\sigma$ is an activation function that is considered as a non-linear mapping (we use leaky ReLu function), and $*$ denotes the 2D convolution. Here $\mathbf{W}$ is of a size $1 \times n \times n \times k$, where 1 is the number of channels in the input (for prestack seicmic data, the value of is always 1), $n$ is the spatial size of a filter, and $k$ is the number of filters. Intuitively, $\mathbf{W}$ applies $k$ convolutions on the input, and each convolution has a kernel size $n \times n$. The output is composed of $k$ feature maps. $\pmb{b}$ is an $k$-dimensional vector, whose each element is associated with a filter. 
The convolution of an $m \times m$ matrix with an $n \times n$ matrix may in fact result in an $(m + n - 1) \times (m + n - 1) $ matrix (full convolution) or in an $(m - n + 1) \times (m - n + 1)$ (valid convolution). 

\begin{figure*}
  \centering
  \includegraphics[width=\textwidth]{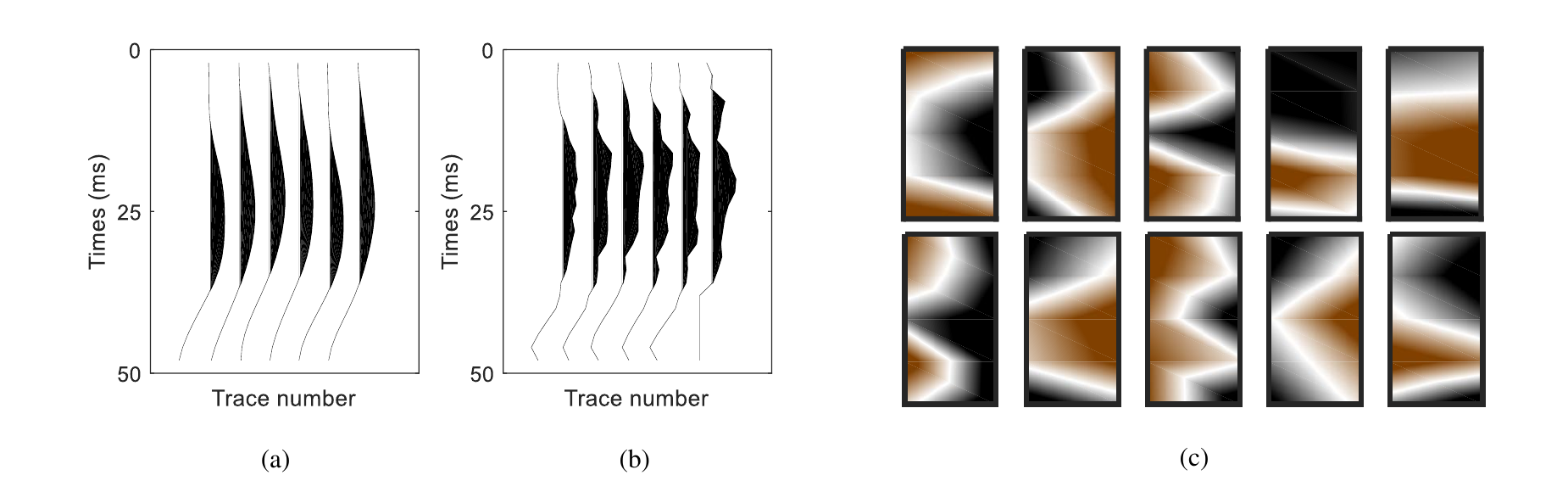}
  \caption{(a) The original waveform of a window of prestack seismic data. (b) The reconstructed prestack seismic data. (c) The extracted features using the convolutional autoencoder network.}
  \label{fig:orf}
\end{figure*}

\textbf{Reconstruction.} In the network, the back two layers are symmetrical with the front two layers, which are called unpooling layer and deconvolution layer. The purpose of this design is to reconstruct the unlabeled input. The reconstruction is obtained using \\
\begin{equation} \label{eq:recon}
  \mathbf{Z}= \sigma \left( \sum_{k\in H}^{} \tilde{\Psi}(\mathbf{Y}_{k}) * \mathbf{\tilde{W}}^{k} + \pmb{c}_{k}\right), 
\end{equation}\\
where again there is one bias $\pmb{c}$ per input channel. $H$ identifies the group of latent feature maps, $\tilde{\mathbf{W}}$ identifies the flip operation over both dimensions of the weights \citep{masci2011stacked}, $\tilde{\Psi}$ donates unpooling operation. 
Generally, max pooling is non-revertible, so instead we formulate a approximate solution that can be used in this case. When reconstructing, a single individual unpooling location is picked randomly and the pooled value is placed there, setting the other unpooling location to zero. This is a good approximation for the max unpooling process, although it may results in some noise in the reconstructed maps. 

\textbf{Training.} We use denoising autoencoding to train the network. The input $\mathbf{X}$ is transformed to some new value $\mathbf{X}_{c}$ by randomly setting values zero with a very small probability and the network is trained to reconstruct the original input. 
Training is done individually layer by layer and posed as an optimization problem. We find weights $\mathbf{W}$ and biases $\pmb{b}$ such that the following loss function is minimized. We use mean squared error (MSE) as the loss function: \\
\begin{equation} \label{MSE}
  L=\frac{1}{2n}\sum_{i=1}^{n} (\mathbf{X}_{i} - \mathbf{Z}_{i})^2.
\end{equation}\\
The weights $\mathbf{W}$ and the biases $\pmb{b}$ are initialized to small random values found using the fan-in and fan-out criteria \citep{hinton2012practical}. 
As for standard networks the back-propagation algorithm is applied to compute the gradient of the error function with respect to the parameters.

We make use of automatic derivatives calculated by TensorFlow with GPUs \citep{abadi2016tensorflow} and iteratively input each data taken from prestack seismic gathers while updating the filters in the direction that minimizes the loss function. An original input and the reconstruction are shown in Figure \ref{fig:orf}-a and \ref{fig:orf}-b. Figure \ref{fig:orf}-c gives the extracted features from the gather.

\subsection{Unsupervised Pattern Recognition}

This is the final stage of our method for generating facies map. The goal is to cluster the one-dimensional vectors, which is derived from the middle feature maps of the convolutional autoencoder network. Clustering can be accomplished by
several unsupervised strategies such as K-means clustering, fuzzy $c$-means, SOM, etc.
Without losing generality, we use K-means algorithm to generate facies map. It is based on minimizing the following function \\
\begin{equation}
  \mathbf{J}_{m}=\sum_{i=1}^{N}\sum_{j=1}^{c}U_{ij}^{m}\left\|x_{i}-C_{j}\right\|^{2},
\end{equation}\\
where $m$ is any real number greater than one, $U_{ij}$ is the degree of
membership of $X_{i}$ in the cluster $j$, $X_{i}$ is the $i^{\text{th}}$ of $d$-dimensional measured data, and $C_{j}$ is the $d$-dimension center of cluster. The cluster centroids are calculated by using the following equation: \\
\begin{equation}
  C_{j}=\frac{\sum_{i=1}^{N}u_{ij}^{m}X_{i}}{\sum_{i=1}^{N}u_{ij}^{m}}.
\end{equation}

\section{Example}

To verify the performance of our method, we applied it to two kinds of prestack data. One is from artificial physical model, another is from LZB region which is a real work area. We selected a interpreted horizon includes known fractures and  
took a window of 48ms along the horizon to obtain data matrices. The learning rate is assigned to 0.02. We set the number of layers of the feature maps to 10, which are obtained by the convolutional filters of size $3 \times 3 \times 10$. On the other hand, we also used poststack data and PCA (retain 90\% of the principal component) based on prestack data to generate facies maps, to which we compared the result derived by the proposed method. All the results are shown in Figure \ref{fig:model_result} and Figure \ref{fig:lzb_result}. 

\subsection{Physical Model}

\begin{figure}
  \vspace{0.2cm}
  \centering
  \includegraphics[width=0.35\textwidth]{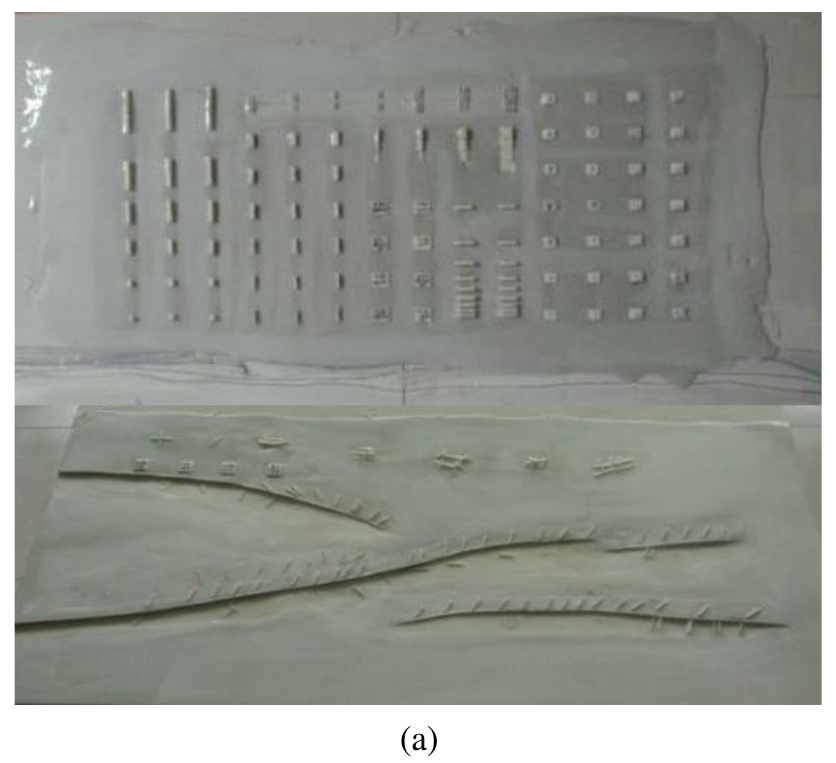}\\
  \includegraphics[width=0.35\textwidth]{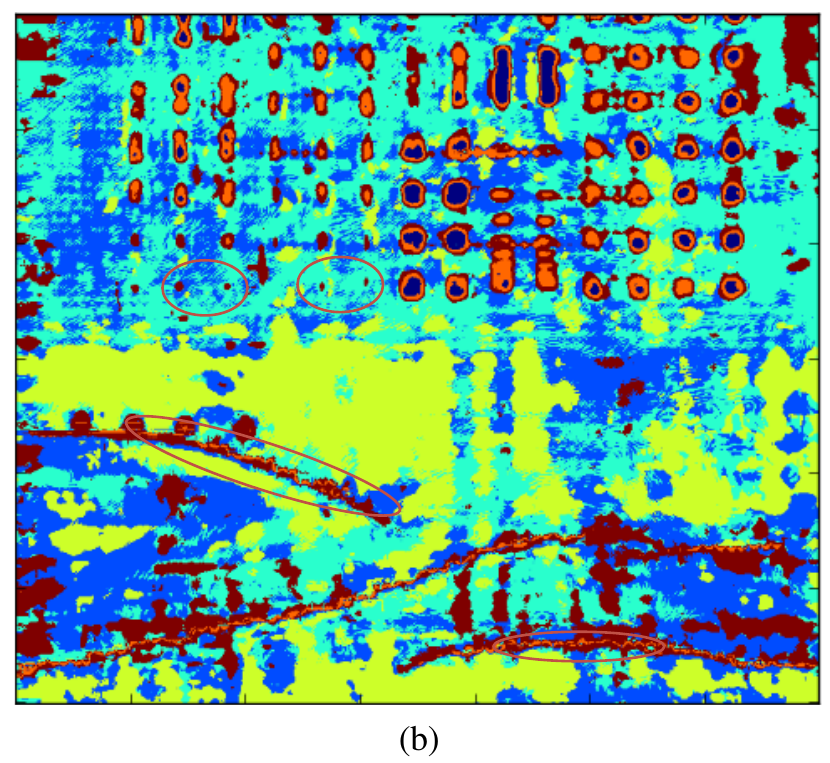}\\
  \includegraphics[width=0.35\textwidth]{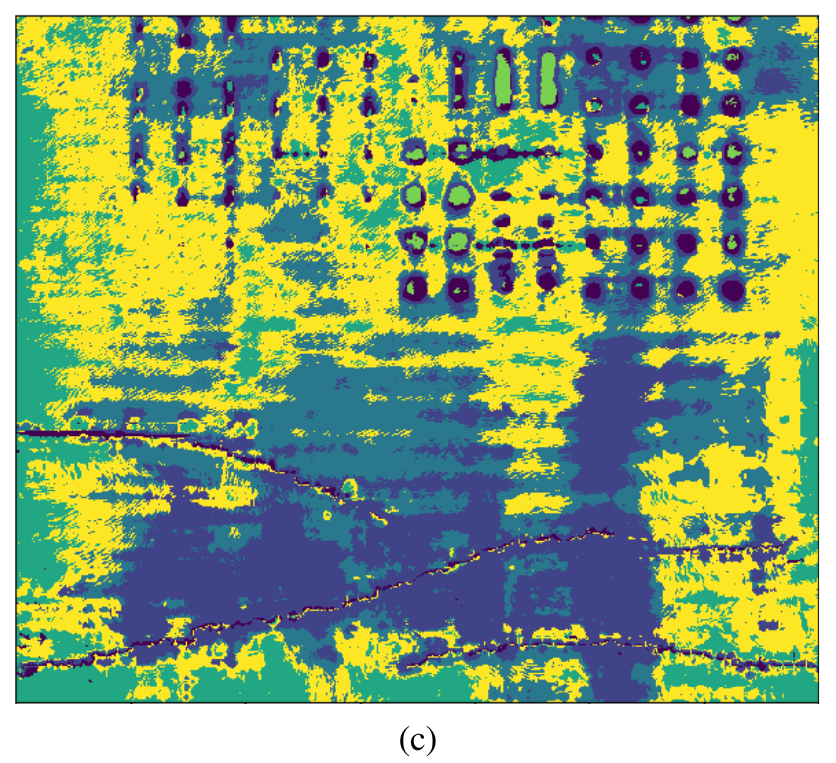}\\
  \includegraphics[width=0.35\textwidth]{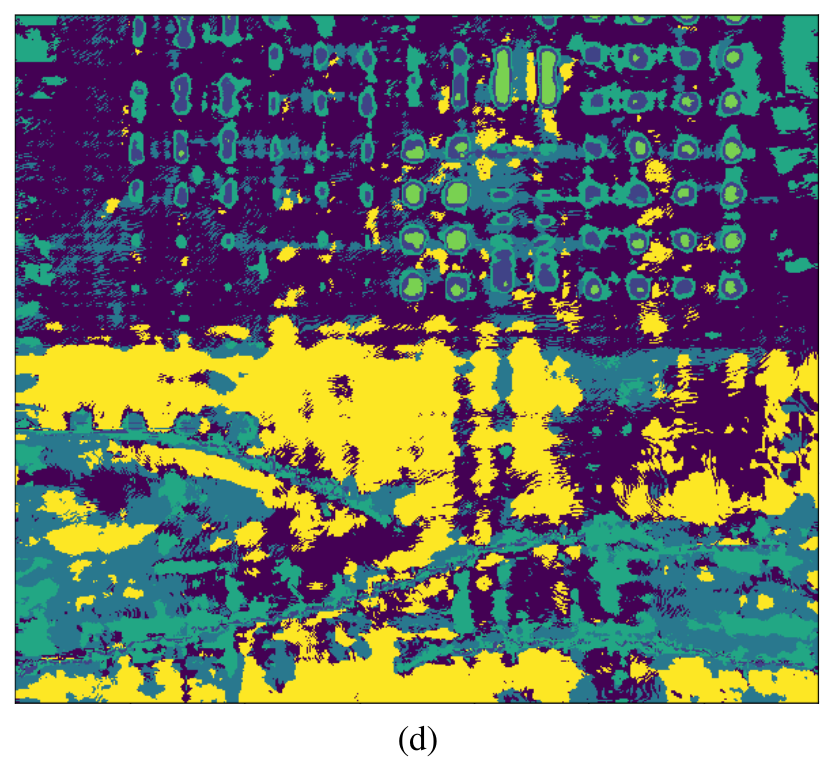}\\
  \caption{Physical model and its facies maps. (a) The prototype of physical model. (b) The result using deep CAE based on prestack data. (c) The result using poststack data. (d) The result using PCA based on prestack data.}
  \label{fig:model_result}
\end{figure}

As shown in Figure \ref{fig:model_result}-a, the physical model consists of three water tanks and many caves. In Figure \ref{fig:model_result}-b, we can clearly see the smallest caves and the river (marked with red circle). However, the small caves are not presented in Figure \ref{fig:model_result}-c. The river in Figure \ref{fig:model_result}-d, is also not as clear as in Figure \ref{fig:model_result}-b.

\subsection{LZB Region}

\begin{figure}[t!]
  \vspace{0.11cm}
  \centering
  \includegraphics[width=0.39\textwidth]{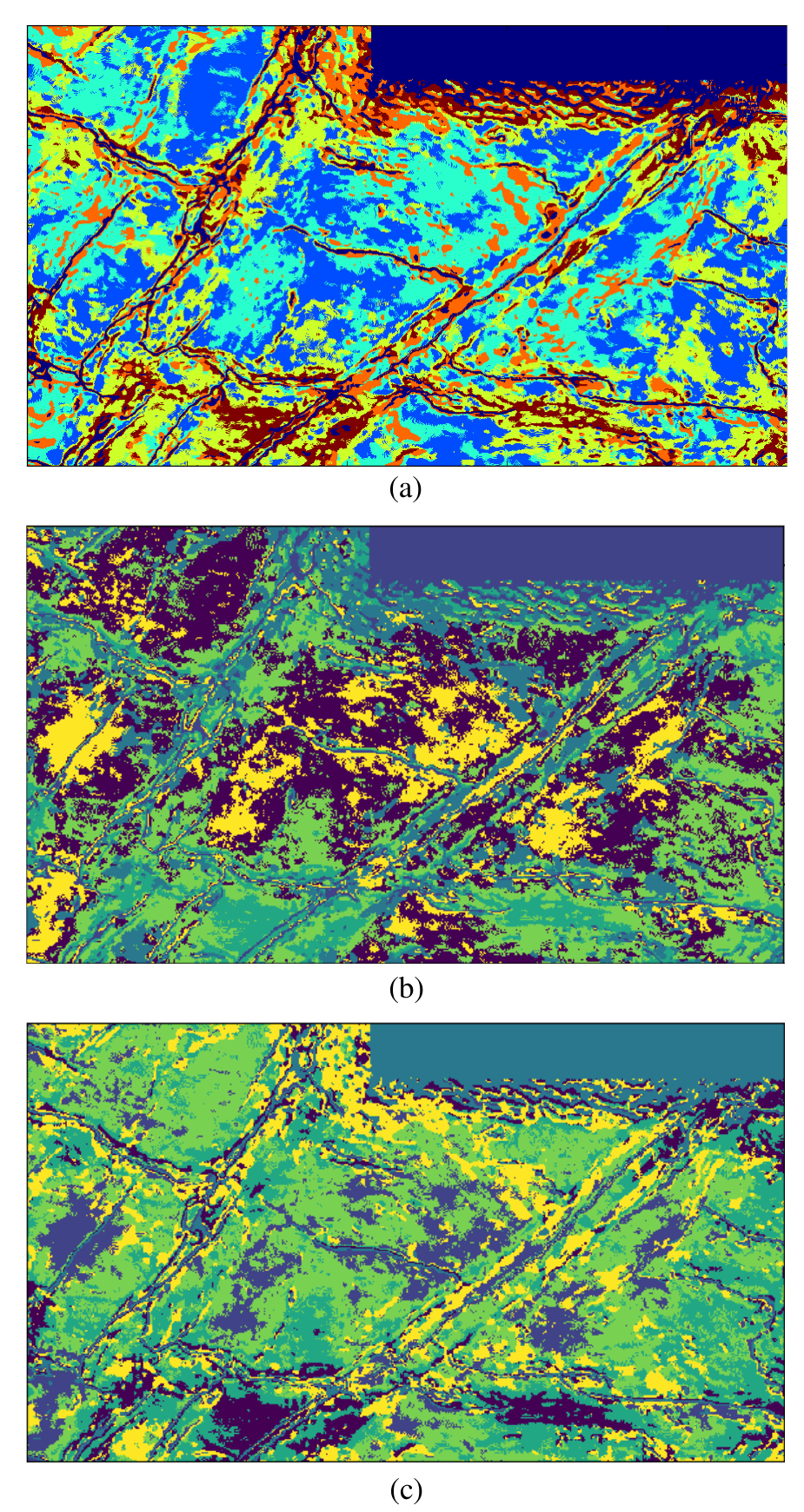}
  \caption{Facies maps of LZB region. (a) The result using deep CAE based on prestack data. (b) The result using poststack data. (c) The result using PCA based on prestack data.}
  \label{fig:lzb_result}
\end{figure}

The facies maps of the three methods are shown in Figure \ref{fig:lzb_result}. Similarly, Figure \ref{fig:lzb_result}-a is generated by our method, of which the effect is better than the other two.

\section{conclusion}

We have presented a novel deep learning approach for seismic facies recognition based on prestack data. We show that with the representation and reconstruction architecture of convolutional neural network and layer-by-layer unsupervised training strategy, reliable features can be learned directly from prestack seismic data. With the learned features we achieved superior performance compared to the current method which is based on poststack data or superficial features learned from prestack data. 
In future work, we will focus on discovering how the parameters, such as the number of levels, the number of hidden units, and the sparsity of active hidden units of the convolutional autoencoders affect the performance.

\section{ACKNOWLEDGMENTS}

This work was supported by the Natural Science Foundation of China (No. U1562218). We 
thank the CNPC Key Laboratory of Geophysical Prospecting under China University of 
Petroleum (Beijing) for their cooperation in providing data and support.

\onecolumn

\bibliographystyle{seg}  


\end{document}